# Improving Crash Frequency Prediction from Simulated Traffic Conflicts Using Machine Learning-Based Microsimulation

Xian Liu [a], Carlo G. Prato [a], Gustav Markkula [b]

[a] School of Civil Engineering, University of Leeds, Leeds, UK

[b] Institute for Transport Studies, University of Leeds, Leeds, UK

Authors Note: Correspondence concerning this paper should be addressed to Xian Liu; Email address: X.Liu2@leeds.ac.uk

**Abstract**

Traffic microsimulation combined with surrogate safety measures has increasingly been used as a proactive alternative to historical crash data for predicting crash frequency for current or planned road infrastructure designs. However, existing microsimulation-based safety studies have adopted simplified rule-based behaviour models, which reproduce traffic flow reasonably well but often fail to generate realistic conflict dynamics, limiting crash prediction accuracy. Recent advances in machine learning (ML)-based behaviour models offer a promising opportunity to potentially improve microsimulation realism and crash frequency predictions by learning human driving behaviour directly from large-scale trajectory datasets. To investigate this possibility, traffic microsimulation was conducted for five real-world signalised intersections in Leeds, UK, using both a standard rule-based model and a state-of-the-art ML model. Simulated vehicle trajectories were analysed using a two-dimensional Time-to-Collision metric to identify simulated conflicts, which were then modelled using Extreme Value Theory to predict crash frequency. Results show that conflicts from the ML model yielded crash predictions in line with the real-world crash data, whereas the rule-based model did not permit meaningful predictions, presumably due to a lack of model calibration to the specific simulated intersections. Directly using ML-generated simulated crashes to predict real-world crash frequency also yielded poor results, suggesting that while current ML models can realistically reproduce conflicts, they are not yet able to generate realistic crashes. Overall, the findings demonstrate that ML-based behaviour models are promising for improving crash frequency prediction from simulated conflicts, without a need for location-specific model calibration, and suggest clear future directions for ML-based traffic microsimulation.

**Key words**: Crash Frequency; Extreme Value Theory; Machine Learning; Surrogate Safety Measures; Traffic Conflicts; Traffic Microsimulation

## 1. Introduction

According to the World Health Organization (WHO, 2023), road traffic crashes remain a leading cause of injury and death worldwide, highlighting the urgent need to improve road traffic safety. A crucial aspect of iteratively improving traffic safety is to accurately assess traffic safety levels, such that traffic safety can be measured before and after interventions, to evaluate their impact. Traditionally, traffic safety assessment has relied on the analysis of historical crash data (Lord & Mannering, 2010). However, crash-based safety assessment is time-consuming, as collecting statistically reliable crash data often takes several years due to the inherently rare and random nature of crashes (Tarko, 2018). In contrast, traffic conflicts, defined as situations where road users approach each other in space and time to such an extent that there is a risk of collision if their movements remain unchanged (Amundsen & Hyden, 1977), occur much more frequently and can be obtained in a significantly shorter period (Arun et al., 2021). Substantial effort has been spent investigating the use of statistical methods such as regression analysis (Lorion & Persaud, 2015; Rajeswaran et al., 2023; Saleem et al., 2014) or Extreme Value Theory (EVT) (Arun et al., 2022; Chen et al., 2024; Hussain et al., 2024) to predict crash frequency from conflicts, and this approach has served as an efficient and proactive alternative to relying purely on historical crash data. To obtain the needed conflict data, there are two main alternative approaches: observing real-world conflicts in the field (Arun et al., 2021; Bhattarai et al., 2023; Zheng & Sayed, 2019) or generating simulated conflicts using traffic microsimulation (Saulino et al., 2015; Wang et al., 2019). However, both of these approaches have clear limitations.

With regards to field observation of real-world conflicts, the observation in itself is nowadays relatively straightforward, thanks to modern automated computer vision techniques which can extract road user trajectories and identify conflicts directly from recorded video data (Fu et al., 2018; Hula et al., 2025). The most important limitation of the field-based approach to conflict measurement is instead that it can only be used to assess safety interventions that have already been implemented in the real world. Consequently, it cannot evaluate the effectiveness of interventions such as infrastructure modifications or technologies (e.g., automated vehicles) that have not yet been deployed, for example to compare multiple alternative designs of the same intervention (Wang et al., 2018). This limitation can significantly slow down the iterative development and optimisation of new road safety interventions.

To overcome the limitations of observing real-world conflicts in the field, traffic microsimulation, a modelling technique that replicates the movement of individual vehicles and their interactions (Toledo et al., 2005), has emerged as an alternative method to simulate road user trajectories and identify simulated conflicts during a simulation period (Arun et al., 2021). This obviates the need for field observations, while also providing a safe and flexible virtual environment for assessing various roadway facilities (Arun et al., 2021; Mahmud et al., 2019; Young et al., 2014). For microsimulation to be a reliable tool, the simulated driving environment should closely replicate real-world traffic, which is heavily dependent on the underlying behaviour models, such as car-following and lane-changing models (Langer et al., 2023; Yan et al., 2023). Nevertheless, previous simulation-based traffic safety studies have been mainly conducted using microscopic simulators like Simulation of Urban Mobility (SUMO) (Krajzewicz, 2010), where the behaviours of background agents are typically governed by relatively simple heuristic rules, such as the Wiedemann model (Wiedemann, 1974) and the Intelligent Driver Model (IDM) (Treiber et al., 2000). These behaviour models were initially developed to simulate reproduce traffic flow dynamics under different traffic conditions, rather than for traffic safety applications (Treiber et al., 2000; Wiedemann, 1974), and are known to have limited realism in simulating conflicts (Guo et al., 2019; Saunier et al., 2007; Tarko & Songchitruksa, 2005), resulting in a substantial fidelity gap between simulated and real-world driving environments.

To enhance the realism of traffic microsimulation, ML-based behaviour models trained on real-world data have been proposed in recent years. ML-based behaviour models aim to create agents that exhibit human-like behaviour (Ransiek et al., 2024). In this context, "human-like" refers not to cognitive modelling of driver decisions, but to the statistical reproduction of trajectory-level interaction patterns observed in real human driving, including variability, timing of manoeuvres, and emergence of conflicts (Ransiek et al., 2024). For example, several studies employed imitation learning to enable simulated vehicles to behave similarly to expert human drivers by directly mimicking observed real-world trajectories (Bhattacharyya et al., 2022; Bhattacharyya et al., 2018; Igl et al., 2022). Reinforcement learning frameworks have also been adopted to refine driving policies through a trial-and-error process, where real-world driving behaviours are translated into reward functions or constraints to ensure both robustness and naturalness (Cornelisse & Vinitsky, 2024; Sackmann et al., 2022; Zhang et al., 2022). In addition, some studies applied autoregressive models that leverage the temporal dependencies in human trajectories, enabling each vehicle to dynamically adjust its next action in response to the historical states of both itself and surrounding traffic (Feng et al., 2022; Philion et al., 2023). Variational-based methods were adopted by some studies to capture the natural diversity of human driving by mapping complex real-world interactions into a simplified mathematical space of different behavioural possibilities (Suo et al., 2021; Tang et al., 2021; Zhang et al., 2023). However, existing applications of these models have mainly focused on their applications in development and evaluation of autonomous vehicles (AVs), with limited attention paid to the potential use of these models for more conventional traffic safety assessment, for example assessment of road infrastructure.

In sum, existing traffic microsimulation safety studies rely on simplified rule-based behaviour models, which struggle to simulate realistic conflicts. Although ML-based traffic microsimulation has recently improved realism, its application to traffic safety assessment remains underexplored, and the present study aims to fill this gap. At the outset of this work, two related but conceptually distinct hypotheses were formulated, corresponding to two different ways of linking microsimulation outputs to real-world crash frequency. The first hypothesis concerns conflict-based crash frequency prediction, where crashes are inferred statistically from simulated traffic conflicts using Extreme Value Theory. Since ML-based traffic microsimulation models are trained on large-scale human driving trajectory datasets that contain abundant instances of near-crash interactions, it was hypothesised that, compared to rule-based behaviour models, ML-based models would generate traffic conflicts that are more consistent with how conflicts arise in human driving. Consequently, conflicts generated by ML-based microsimulation were expected to yield more accurate predictions of real-world crash frequency than conflicts generated by a rule-based model.

The second hypothesis concerns crash-based crash frequency prediction, where crash frequency is estimated directly from crashes generated by the microsimulation model itself. Unlike most rule-based models, which are explicitly designed to avoid collisions and therefore do not generate crashes (Van Lint & Calvert, 2018), ML-based models can produce simulated crashes (Cornelisse & Vinitsky, 2024). However, despite their improved behavioural realism, crashes remain extremely rare in even the largest real-world driving datasets used for training ML models (Liu & Feng, 2024; Yan et al., 2023). Therefore, the second hypothesis of this study was that ML-simulated crashes would not reliably reproduce the mechanisms by which human driving behaviour leads to real-world crashes, and thus would not provide accurate predictions of observed crash frequency. Accordingly, the contribution of this study lies in a comparative evaluation of different behavioural modelling approaches for crash frequency prediction, rather than in developing a transferable crash prediction model.

The remainder of the paper is organized as follows: Section 2 presents the study methodology, Section 3 reports the analysis results, which are discussed in Section 4. Finally, Section 5 concludes the paper.

## 2. Methodology

### *2.1. Field data collection*

For consistency, all of the intersections studied here were four-arm signalized intersections; this type of intersection involves diverse vehicle interactions and is widely studied in traffic safety research (Ali et al., 2023). The study was intentionally conducted using five signalised intersections (Fig. 1) to balance methodological rigour, computational feasibility, and comparability with prior EVT-based crash prediction studies. EVT–based conflict analysis is typically applied at a limited number of sites, with previous studies using between two (Arun et al., 2022) and ten locations (Wang et al., 2018), depending on data availability and study objectives. In the present work, the primary objective is not to develop a transferable crash prediction model, but to perform a controlled comparative evaluation of crash frequency prediction performance between rule-based and ML-based traffic microsimulation models under identical conditions. For this purpose, a smaller number of carefully selected intersections with heterogeneous crash frequencies is sufficient to assess relative model performance. Using a larger number of locations would substantially increase computational demands (particularly for ML-based microsimulation) without strengthening the comparative conclusions of the study.

To evaluate the consistency between simulated and observed traffic volumes, traffic volume data for each intersection during weekday morning peak hours (8:00–9:00 AM) were collected using the Leeds City Council's (LCC) VivaCity AI-powered computer vision sensors (VivaCity, 2025). For each intersection, data on police-reported vehicle-to-vehicle crashes (property damage only, slight injury, serious injury, and fatal crashes) occurring within 20 m of the intersection, as defined in the STATS19 reporting guidelines (Transport, 2022), during the period 2021–2023 were obtained from LCC. The actual annual crash frequency for each intersection was then calculated from these data as the ground truth for crash prediction. Additionally, Poisson confidence intervals for these ground truth crash frequencies were estimated, as Songchitruksa and Tarko (2006):

$$\lambda : \frac{1}{2n}\chi^2_{2y_i,1-\alpha/2} \le \lambda \le \frac{1}{2n}\chi^2_{2(y_i+1),\alpha/2} \quad (1)$$

where $\lambda$ is the annual crash frequency, $\chi^2$ is the Chi-squared distribution, $y_i$ is the total observed crash counts for intersection $i$ for $n$ years, $\alpha$ is the significance level, set to 0.05 in current study (Songchitruksa & Tarko, 2006), corresponding to a 95% confidence interval.

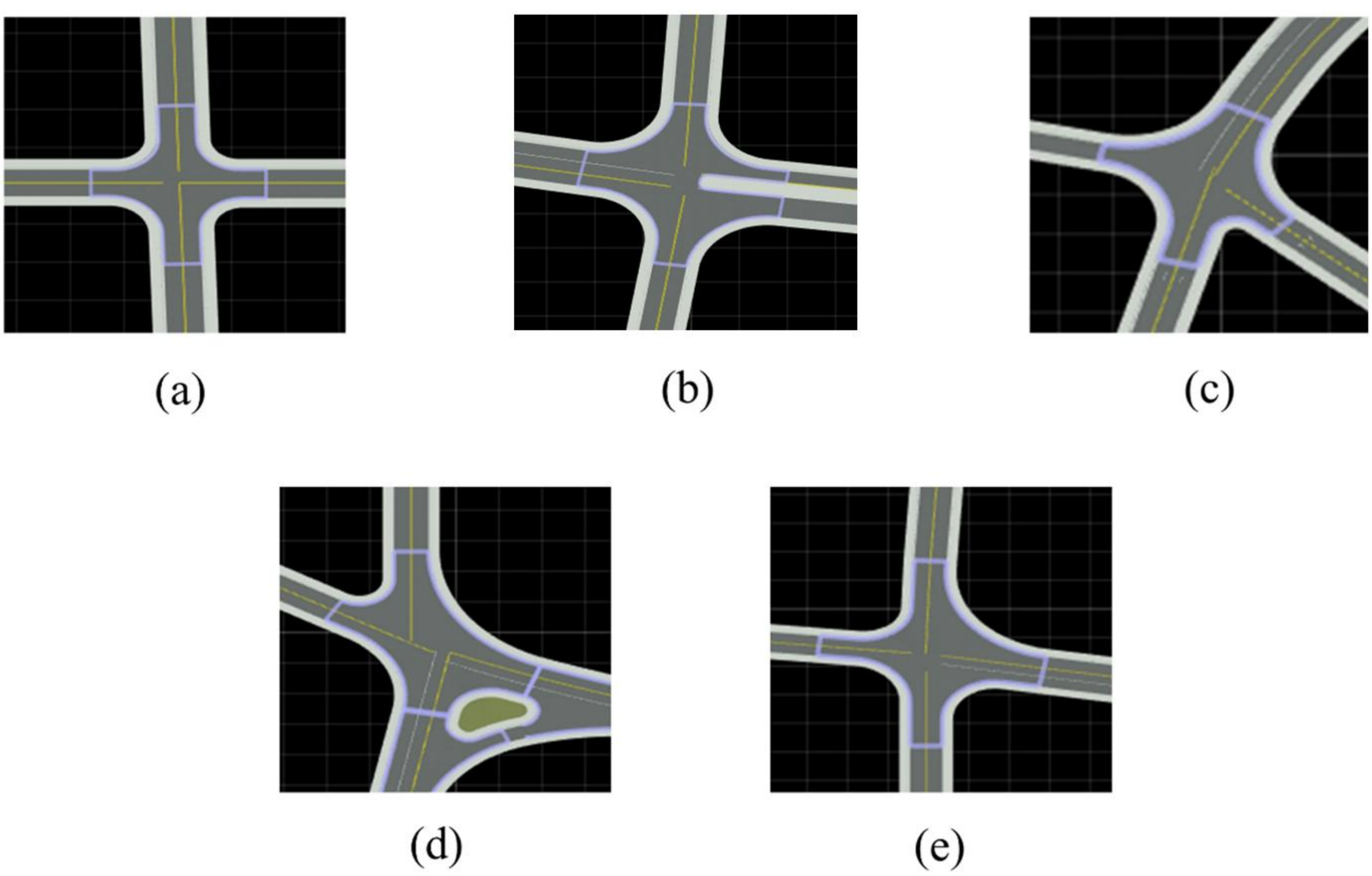

**Fig. 1. Layouts of the five studied intersections. The solid lines across the lanes denote the stop lines for traffic lights in the intersections.**

*2.2. Traffic microsimulation*

*2.2.1. Rule-based traffic microsimulation*

This study used the open-source software SUMO as the microscopic simulator for rule-based traffic microsimulation (Krajzewicz, 2010). SUMO natively includes a number of rule-based models. In this study, the Wiedemann model was chosen as the main rule-based model under study, since this was the model used in past EVT-based crash prediction work (Wang et al., 2018). To simulate the Leeds intersections in SUMO, the corresponding maps were first downloaded from OpenStreetMap (OSM) in .osm format and then converted into SUMO-compatible .net.xml format using the built-in netconvert tool. The default SUMO parameters for the Wiedemann model were used, and the simulated traffic volumes were calibrated to match the collected real traffic volume data. For each intersection, the simulation length was set to 1 hour, consistent with previous studies using traffic conflict simulation to predict crash frequencies (Lorion & Persaud, 2015; Rajeswaran et al., 2023; Wang et al., 2018), and repeated five times. Repeating simulations is a common practice in traffic simulation studies to reduce random variability (Rajeswaran et al., 2023; Saleem et al., 2014; Saulino et al., 2015; Wang et al., 2018). The simulation step size was set to 0.1 s. Upon completion, the vehicle trajectory data were extracted for further analysis.

### *2.2.2. ML-based traffic microsimulation*

For the ML-based traffic microsimulation, this study employed the TorchDriveSim microscopic simulator developed by Inverted AI (Scibior et al., 2021). The simulator provides a deep generative model, Control-ITRA (Lioutas et al., 2025). Control-ITRA is based on the ITRA architecture, originally presented in (Scibior et al., 2021), and subsequently trained and updated on a large and globally diverse dataset (exceeding 1000 hours of traffic data, collected across 19 countries) (Lioutas et al., 2025), enabling realistic, human-like driving behaviours across a wide range of potential traffic scenarios. Similar to SUMO, TorchDriveSim can also be used in conjunction with real-world maps for simulation studies. The OpenDRIVE format representations of the intersections were converted into the Lanelet2 format compatible with the TorchDriveSim simulator, using the Inverted AI map-converter tool. The simulated representation of each intersection, i.e., road width, road length, speed limits, and signal timing, was consistent across the SUMO and TorchDriveSim simulations.

The ML-based microsimulation was implemented using two APIs provided by Inverted AI: INITIALIZE and DRIVE. INITIALIZE is used to spawn vehicles in a given scene, while DRIVE is used to advance the simulation by one time step, updating the states of all vehicles in the scene. Consistent with the SUMO simulations, each intersection was simulated for 1 hour and repeated five times, using a simulation step of 0.1 s. The Inverted AI ML model is optimized for simulation of scenarios lasting up to 20 s, and the number of vehicles in the simulation is limited, with the maximum allowed varying by location (Inverted AI, 2022), but this maximum vehicle limit did not prevent the target traffic volume from being achieved at the studied intersections. To achieve the same 1-hour simulation duration as with the Wiedemann model, the ML simulation was divided into 360 segments of 10s each. This segmentation did not affect conflict identification, which was performed continuously across segments. It is worth noting that the ML-based microsimulation is computationally demanding. 1-hour simulation of a single intersection takes around 10 hours and repeating it five times for an intersection thus requires approximately 50 hours. Similar to the SUMO simulations, trajectory data were extracted upon completion.

To assess the consistency between simulated and observed traffic volumes, the GEH statistic was employed (Caliendo & Guida, 2012). This standard measure compares actual and simulated volumes and is widely used for traffic model validation. The GEH statistic is calculated as follows:

$$GEH = \sqrt{\frac{2(M-C)^2}{M+C}} \tag{2}$$

where $M$ is the actual hourly traffic volume, and $C$ is the averaged simulated hourly traffic volume, calculated separately for the Wiedemann- and ML-based simulations.

*2.3. Conflict identification*

This study employed a two-dimensional TTC metric to identify potential conflicts (Jiao, 2023). Traditional one-dimensional TTC is commonly used to estimate the time remaining before a potential collision between two vehicles (Hayward, 1972), and works well for rear-end conflicts on straight roads. However, at urban intersections, vehicles frequently perform lateral manoeuvres such as lane changes or turning, and their paths are often curved or non-linear (Ward et al., 2015). As a result, the standard TTC calculation is not sufficient for identifying conflicts in such settings. To address this limitation, the two-dimensional TTC framework from (Jiao, 2023) was adopted, briefly described here:

At each time step, Distance-to-Collision (DTC) is computed as the shortest distance between the vehicles' bounding boxes along their relative velocity vector. Specifically, given the corner points $C_i$ of the target vehicle $i$ and the corner points of $C_j$ the interacting vehicle $j$, a line can be drawn from each corner point $c_i \in C_i$ in the direction of the relative velocity vector $v_{ij} = v_i - v_j$. Any intersection points between the line and the edges of vehicle $j$'s bounding box are recorded as $k_j$. Distances between $c_i$ and the corresponding $k_j$ are calculated:

$$d_{i \to j} = \begin{cases} \| k_j - c_i \|, & (k_j - c_i)v_{ij} \geq 0 \\ \infty, & (k_j - c_i)v_{ij} < 0 \text{ or } k_j \text{ does not exist} \end{cases} \tag{3}$$

Considering each of vehicles $i$ and $j$ in turn as the target, two distance sets $D_{i\to j}$ and $D_{j\to i}$ are obtained, from which the minimum bounding-box distance along their relative velocity can be identified:

$$DTC = \min\{\min\{D_{i\to j}\}, \min\{D_{j\to i}\}\} \tag{4}$$

Finally, the TTC can be computed as:

$$TTC = \begin{cases} \dfrac{DTC}{\| v_{ij} \|}, & DTC \neq \inf \\ \infty, & DTC = \inf \end{cases} \tag{5}$$

Subsequently, the minimum TTC value for each conflict event was extracted and incorporated into the Extreme Value Theory (EVT) model for further analysis (Guo et al., 2021; Zheng et al., 2019).

*2.4. Extreme Value Theory analysis*

After identifying conflicts, the next step involves quantifying the statistical relationship between traffic conflicts and crash frequency. Traditionally, a common approach has been to use regression analysis (Caliendo & Guida, 2012; Lorion & Persaud, 2015; Rajeswaran et al., 2023). However, this regression analysis still relies on crash data, which are similarly limited in availability as in traditional crash-based safety assessment (Farah & Azevedo, 2017). In contrast, EVT-based methods infer unobserved crash probabilities from the tail behaviour of observed conflicts, without relying on crash records. EVT-based methods have been increasingly adopted in traffic safety analysis and have demonstrated higher accuracy in predicting crash frequency than regression models (Ali et al., 2023; Wang et al., 2018). Therefore, EVT was used in this study to predict crash frequency. There are two main approaches within the EVT context: block maxima (BM) and peak over threshold (POT) (Coles et al., 2001). Previous research applying EVT to conflict-based crash frequency prediction indicates that the POT method is more effective than the BM method (Tarko, 2018; Wang et al., 2019; Zheng et al., 2014). Therefore, a POT-based EVT framework was employed in this work.

In the POT approach, observations with measurements exceeding a predetermined threshold are considered extremes (Coles et al., 2001); in the context of traffic safety assessment the extremes of interest are the actual crashes. Let $\{X_1, X_2, ..., X_n\}$ be independent and identically distributed random variables. Over a threshold $u$, the conditional distribution of exceedances $y = X - u$ is defined as $F_u(y) = \Pr(X - u \le y \mid X > u)$. With a large enough $u$, the conditional distribution of exceedances is well approximated by a Generalized Pareto (GP) Distribution. The distribution function expressed as follows:

$$G(y) = 1 - (1 + \xi \frac{y}{\sigma})^{-1/\xi} \tag{6}$$

where $\sigma > 0$ is the scale parameter, and $-\infty < \xi < +\infty$ is the shape parameter. The two parameters are estimated using the Maximum Likelihood Estimation (MLE) method (Zheng et al., 2014).

The tail of an extreme value distribution is particularly important, as EVT allows extrapolating beyond the range of observed data. In road safety applications, this involves using frequently observed traffic conflicts to estimate the probability of rarer crashes. By fitting a GP distribution to the negated TTC values of conflicts, where smaller TTC indicates higher crash risk and the extreme value of TTC $\leqslant 0$ corresponds to an actual crash, the probability of a crash can be estimated from the tail of the fitted distribution (Bhattarai et al., 2023; Guo et al., 2021; Wang et al., 2018):

$$R = \Pr\{z \ge 0\} = 1 - G(0) = (1 + \xi \frac{(0 - u)}{\sigma})^{-1/\xi} \tag{7}$$

where $R$ is the crash risk, $z$ is the is the maximum negated TTC, and $G$ is the extreme value distribution.

The predicted crash frequency over a period of time $T$ can then be obtained as:

$$N = R \times \frac{T}{t} \tag{8}$$

where $N$ is the predicted annual crash frequency, $T$ is the total number of hours in a year, $t$ is the observation period at each intersection.

To quantify the accuracy of crash frequency predictions, we used three different metrics: (1) hit rate, defined as in the percentage of predicted values that fall within the Poisson confidence intervals of the actual annual crash frequency (Songchitruksa & Tarko, 2006), (2) mean absolute error (MAE), which quantifies the average magnitude of the errors between predicted and actual crash frequency, providing a straightforward measure of overall deviation (Chai & Draxler, 2014), and (3) root mean square error (RMSE), which calculates the square root of the mean squared differences, giving greater weight to larger discrepancies (Chai & Draxler, 2014). Together, these metrics allow for a quantitative comparison of model performance and the assessment of prediction accuracy.

## 3. Results

### *3.1. Traffic volume simulation validation*

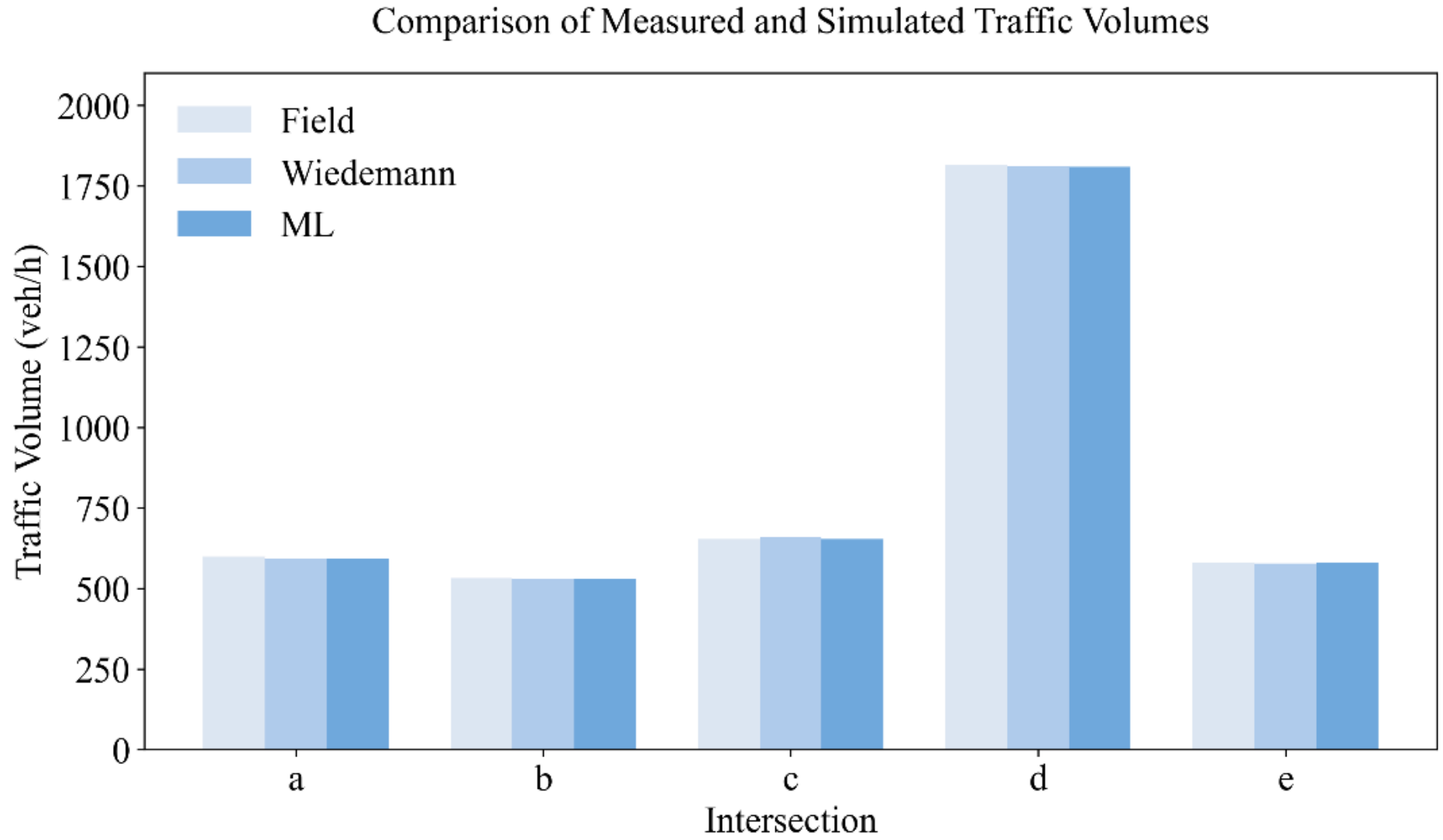

**Fig. 2. Comparison of field and simulated traffic volumes**

Fig. 2 presents a comparison between the field and simulated traffic volumes. For all intersections, the GEH values calculated using Eq. (1) for the traffic volumes simulated by both the Wiedemann and ML models are below the recommended value of 5 (Caliendo & Guida, 2012), indicating a good level of agreement between the simulated and real traffic volumes.

### *3.2. High-level comparison of model behaviour*

To give an impression of what the behaviour generated by the Wiedemann and ML models looks like, this section shows example vehicle trajectories, and an analysis of the distributions of TTC, linear speed, and angular speed.

#### *3.2.1. Trajectories*

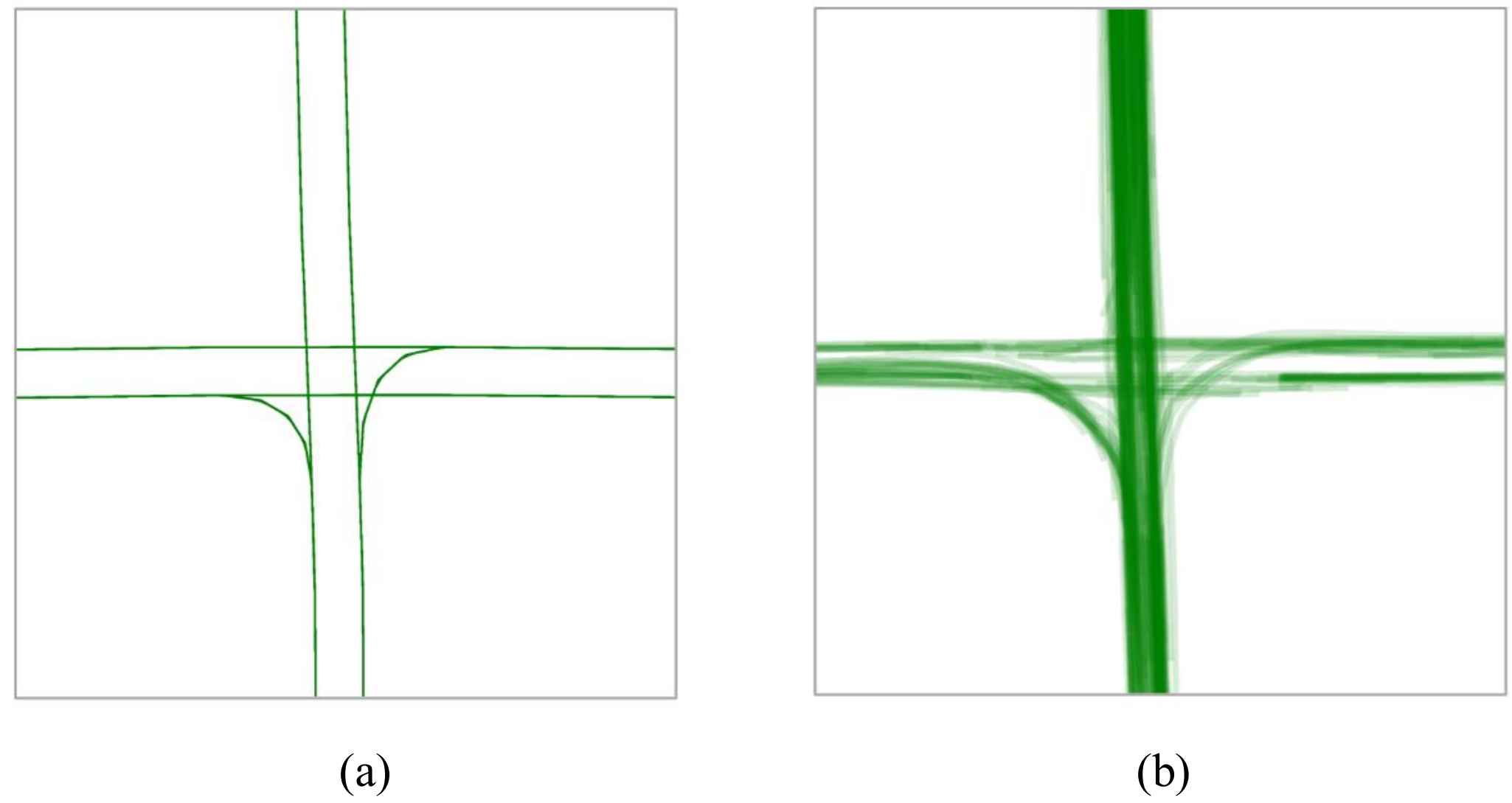

(a) (b)

**Fig. 3. Example trajectories from the (a) Wiedemann and (b) ML models in intersection a**

Fig. 3 shows an example of 500 random trajectories simulated from (a) Wiedemann and (b) ML models in intersection a. The trajectories generated by the Wiedemann appear more rigid and robotic: vehicles follow nearly identical paths. In contrast, the trajectories simulated by the ML model exhibit considerable variability among vehicles even when following similar routes, and appear qualitatively more similar to typical human driving trajectories, as observed in the real-world data reported by Saldivar-Carranza and Bullock (2024).

*3.2.2. Distribution of TTC*

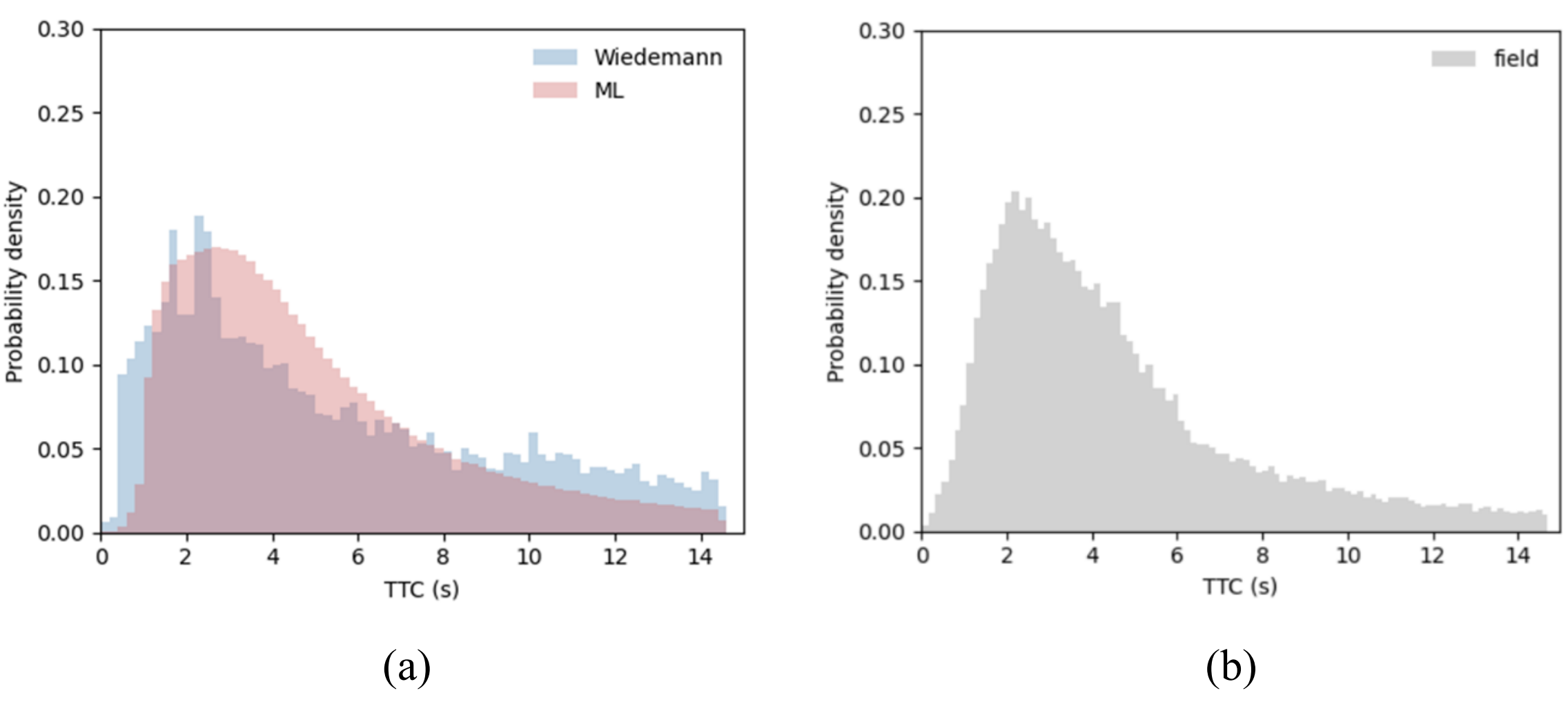

(a) (b)

**Fig. 4. TTC distributions of (a) Wiedemann and ML models and (b) field data**

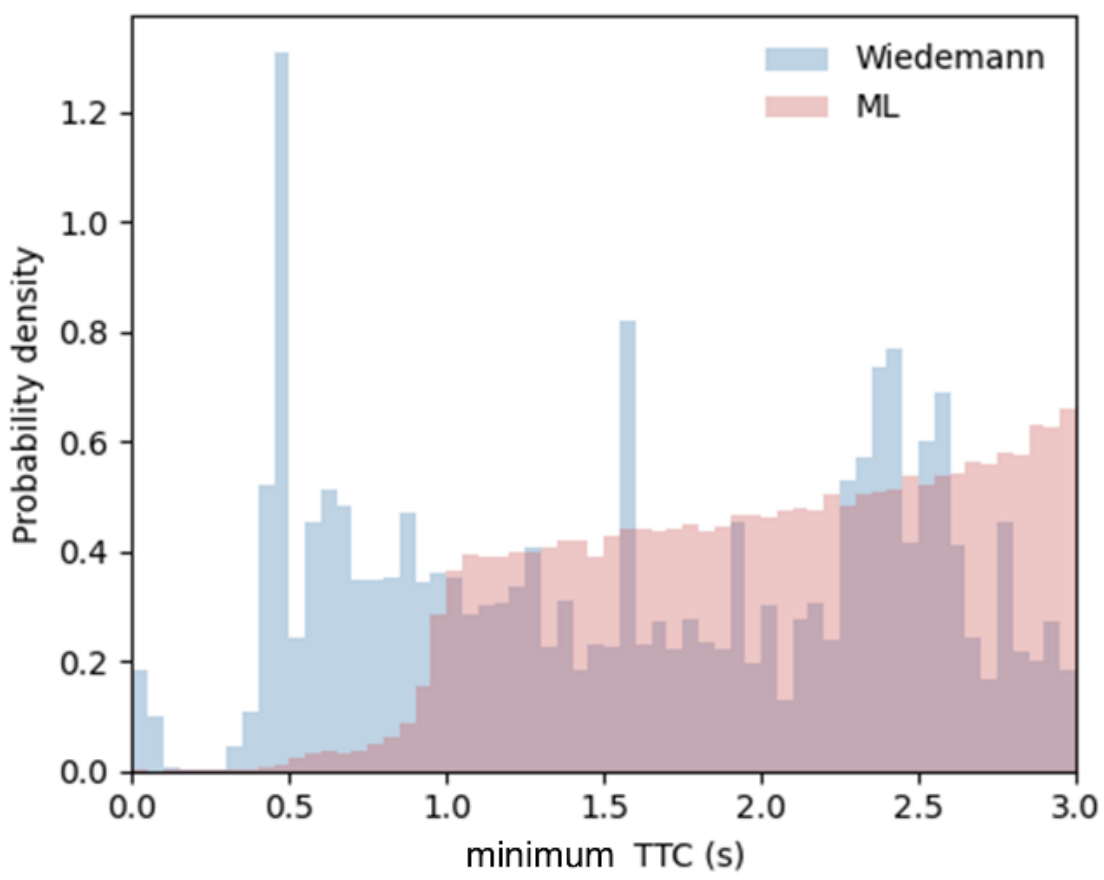

**Fig. 5. Minimum TTC distributions of Wiedemann and ML models**

Fig. 4 shows the distributions of two-dimensional TTC obtained from the Wiedemann and ML simulated trajectories, as well as TTC values derived from field observations at signalized intersections (Lin et al., 2025). To quantitatively compare the TTC distributions, the KL-divergence metric is employed (Kullback, 1997). KL-divergence ranges from 0 to infinity, with smaller values indicating greater similarity between the two distributions. Notably, the KL-divergence between the ML model and the field data (0.025) is lower than that between the Wiedemann model and the field data (0.106), indicating that the distributions generated by the ML model align more closely with the field data. Furthermore, Fig. 5 presents the minimum TTC distributions of the Wiedemann and ML models. The Wiedemann model exhibits discontinuity and multi-modality, while the ML model generates a more continuous and smooth distribution.

*3.2.3. Conflict heat map*

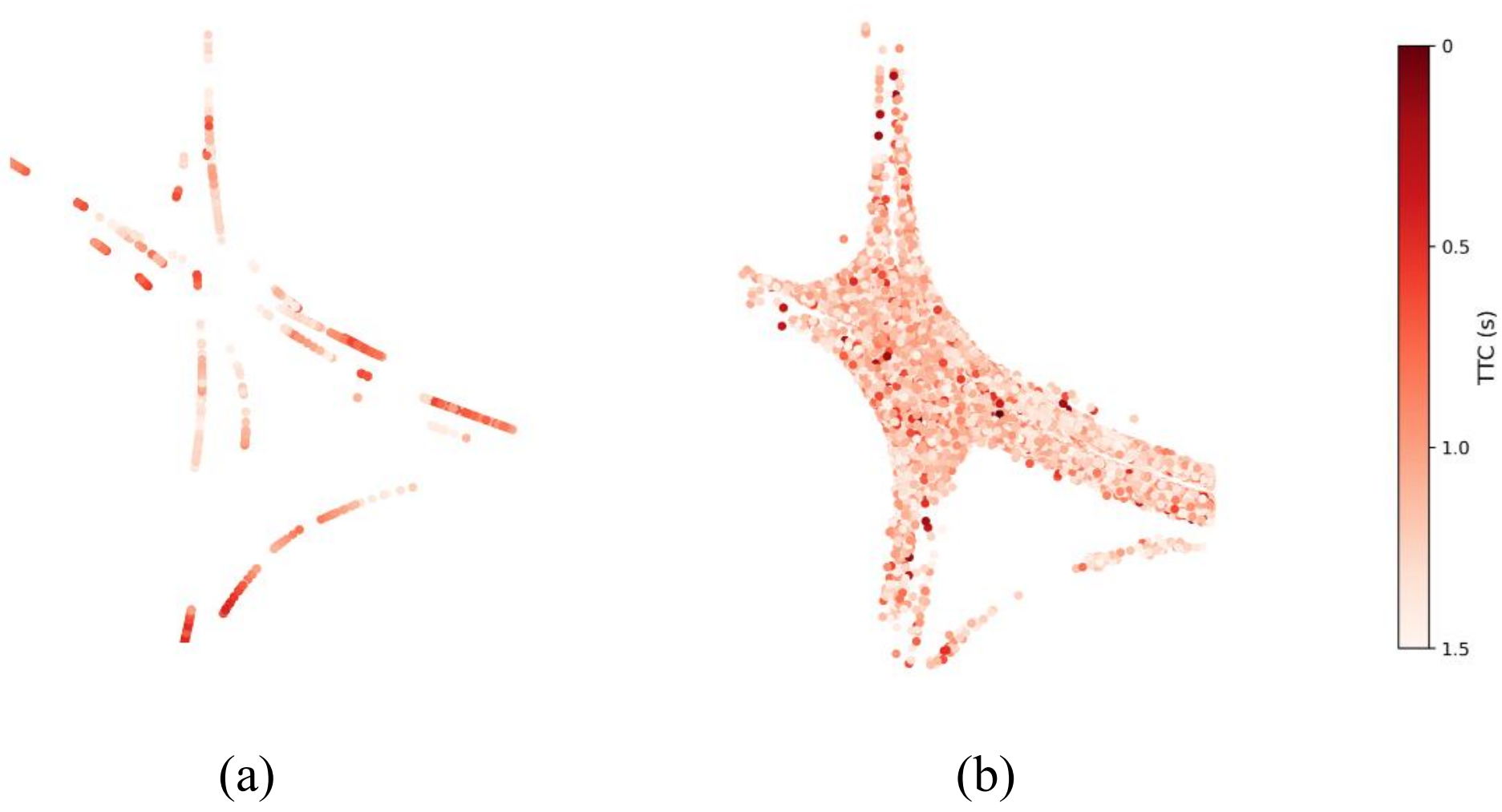

(a) (b)

**Fig. 6. Example conflict heatmap from the (a) Wiedemann and (b) ML models in intersection d**

Fig. 6 shows example conflict heatmaps for intersection d generated by the (a) Wiedemann and (b) ML models in intersection d. Only conflicts with minimum TTC $\leqslant$ 1.5 s are displayed, following the commonly adopted TTC threshold for safety-critical conflicts (Sayed et al., 1994). As shown in Fig. 6(a), the conflicts heatmap produced by the Wiedemann model are highly discrete, sparse, and tightly constrained along narrow, linear paths. In contrast, the ML model (Fig. 6(b)) exhibits a denser cloud-like heatmap of conflict points, and appear more similar to conflict heatmap generated by human drivers, as observed in the real-world data reported by Sarkar et al. (2025).

*3.2.4. Distribution of linear speed*

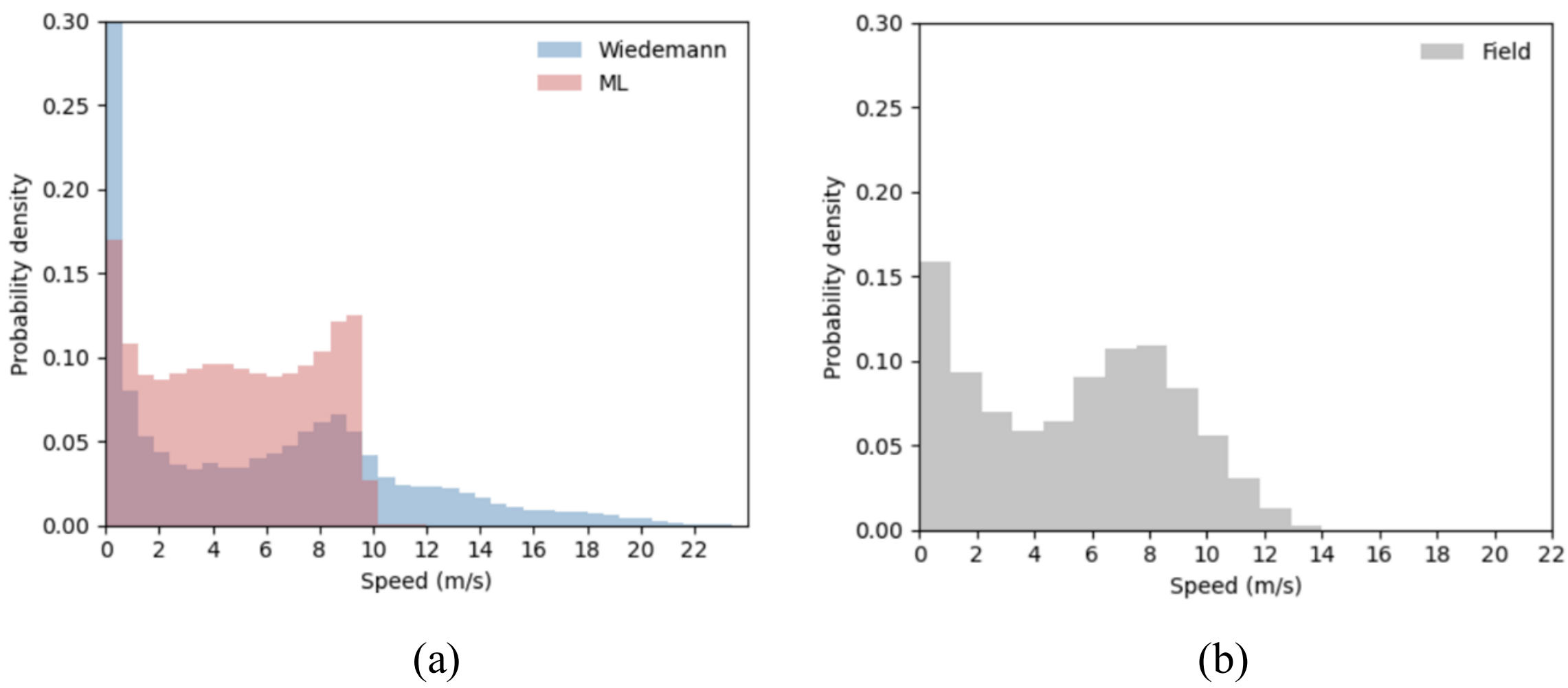

**Fig. 7. Linear speed distributions of (a) Wiedemann and ML models and (b) field data**

Fig. 7 compares the distributions of linear speed obtained from the Wiedemann and ML simulations, with the distribution derived from field observations at signalized intersections (Yan et al., 2023). The KL-divergence between the ML model and the field data (0.096) is lower than that between the Wiedemann model and the field data (0.292), suggesting a greater similarity between the distribution produced by the ML model and the field data. Additionally, as shown in Fig. 7, the ML model qualitatively generates a speed distribution comparable to the field data. By contrast, the Wiedemann model exhibits a noticeably longer tail of higher speeds than the field data. However, one limitation of the ML model is its lack of speeds above 10 m/s, which may be due to the fact that the dataset used to train the ML model contains relatively few instances of speeds in this range (Lioutas et al., 2025).

*3.2.5. Distribution of angular speed*

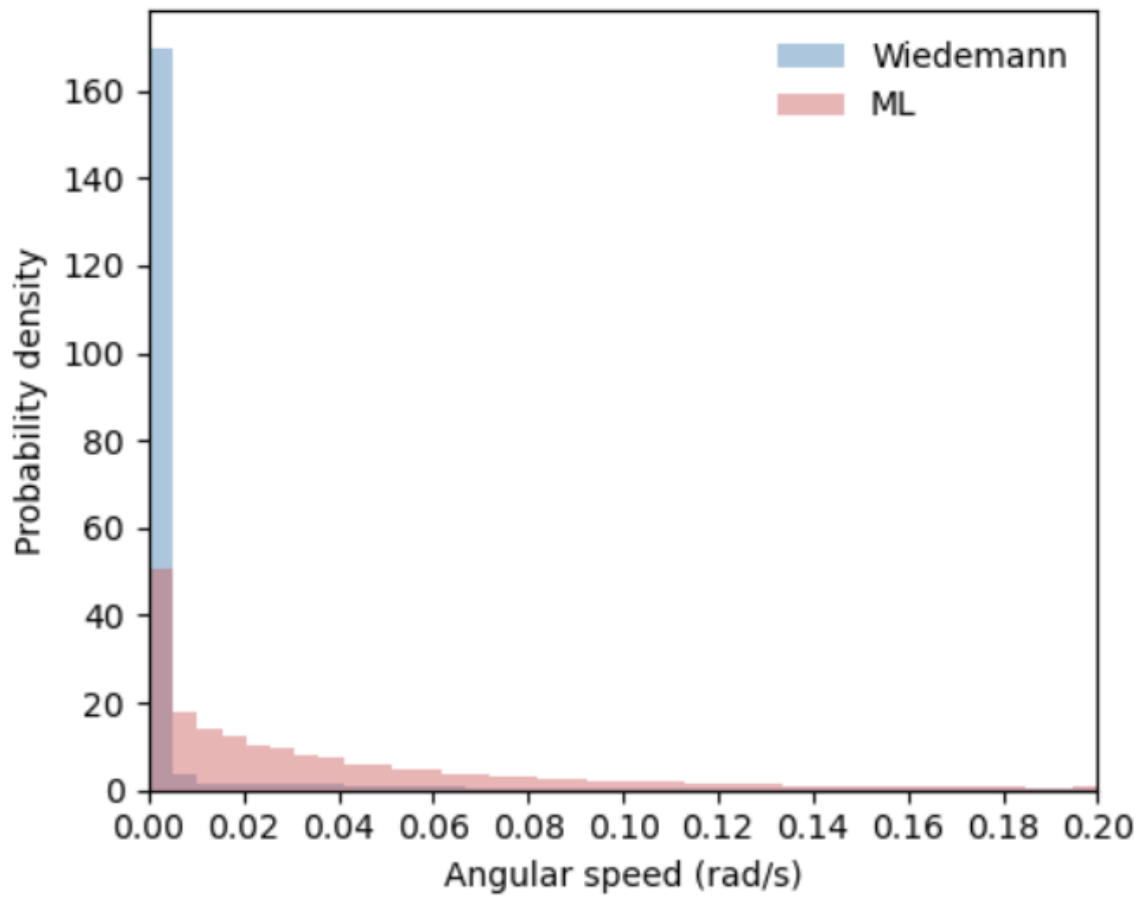

**Fig. 8. Angular speed distributions of Wiedemann and ML models**

Fig. 8 compares the distributions of angular speed obtained from the Wiedemann and ML models, showing that the Wiedemann model produces a substantially higher proportion of zero angular speed than the ML model (76.83% vs. 12.49%). This indicates that the simulated movements in the Wiedemann model are highly regular and stereotyped, as was clear also from Fig. 3. The trajectories generated by the ML model exhibit more small fluctuations in heading direction, akin to imperfect driver control (Ding et al., 2022), and arguably more human-like.

*3.3. Crash frequency prediction and comparison*

EVT models were fitted to the model-generated conflicts to predict crash frequency. To more comprehensively compare the crash frequency prediction capabilities of the Wiedemann and ML models, two fitting approaches were applied to each intersection and each behaviour model. The first was a non-pooling approach, where EVT models were fitted separately to the conflicts obtained from each simulation repeat. The second was a pooling approach, where conflicts obtained from the five simulation repeats were aggregated into a pooled dataset, to which EVT models were then fitted. Before each fit, an appropriate conflict indicator threshold needs to be identified separately. In the current study, the mean residual life plot and the threshold stability plots for the modified scale and shape parameters were employed to determine the threshold, in line with previous authors (Farah & Azevedo, 2017; Guo et al., 2021; Wang et al., 2018). A negated TTC threshold can be identified by observing the point at which the mean residual life plot exhibits a linear trend and the modified scale and shape parameters remain stable (Zheng et al., 2014).

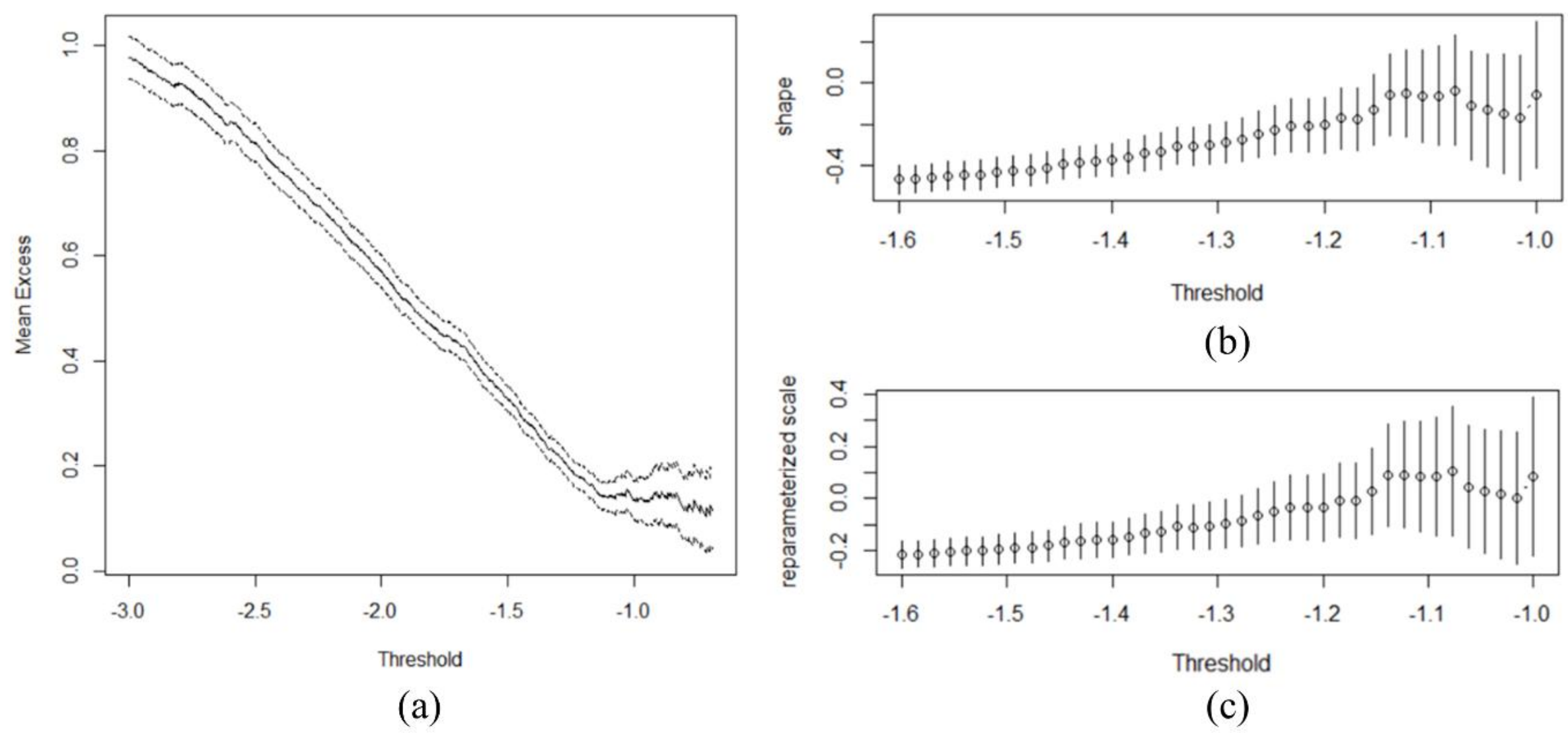

**Fig. 9. (a) Mean residual life plot (b) Modified Scale Parameter Stability Plot and (c) Shape Parameter Stability Plot for the negated TTC margin**

Figure. 9 illustrates an example of selecting the appropriate negated TTC threshold for a single simulation. As shown in Fig. 9 (a), the mean residual life is approximately linear when the negated TTC is less than −1.1, while both the modified scale and shape parameters exhibit stability around −1.1, as shown in Fig. 9 (b) and (c). Therefore, −1.1 was selected as the threshold for this example.

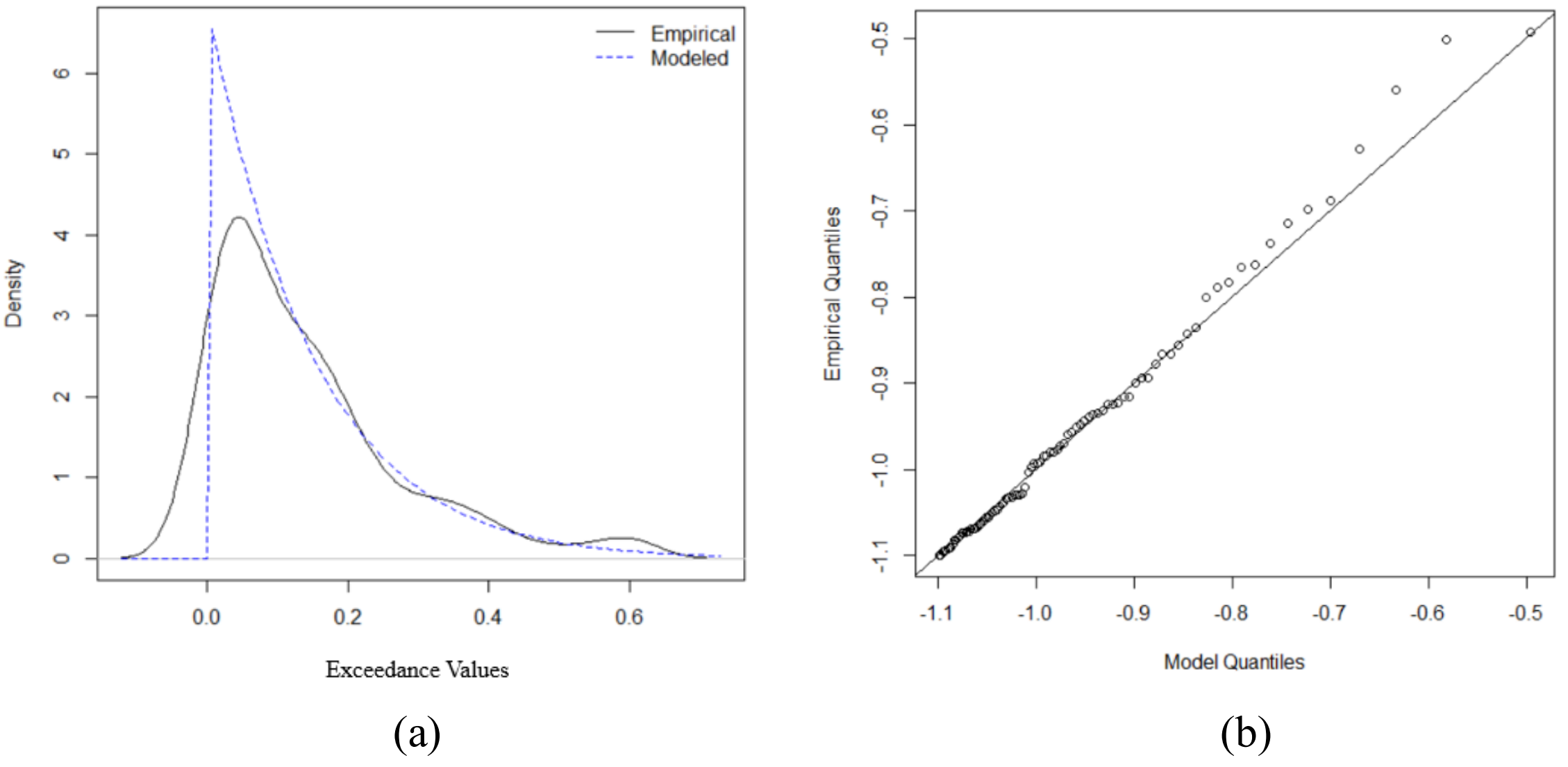

**Fig. 10. (a) Kernel probability density function plot (b) simulated QQ plot**

Fig. 10 shows, for the same example as in Fig. 9, the kernel probability density functions of the empirical and modelled exceedance values, along with the corresponding simulated QQ plot. Both plots indicate a good fit between the modelled GP distribution and the empirical data (Farah & Azevedo, 2017; Wang et al., 2018). The same procedure was applied separately to the conflicts from the Wiedemann and ML models for each intersection and each fitting approach. Tables 1 and 2 present the thresholds and parameter estimates under both non-pooling and pooling approaches.

**Table 1**

**Thresholds and parameter estimates under the non-pooling approach.**

| | Wiedemann model | | | ML model | | |
|---|---|---|---|---|---|---|
| Intersection No. | Thresholds | $\sigma$ | $\xi$ | Thresholds | $\sigma$ | $\xi$ |
| a | [-1.50, -0.80] | [1.42, 4.29] | [-3.79, -3.09] | [-2.00, -1.60] | [0.57, 0.99] | [-0.49, -0.30] |
| b | [-1.40, -0.90] | [0.92, 1.86] | [-1.33, -1.03] | [-1.60, -1.30] | [0.35, 0.60] | [-0.53, -0.29] |
| c | [-1.50, -1.00] | [0.52, 1.22] | [-1.12, -0.79] | [-1.60, -1.10] | [0.15, 0.50] | [-0.42, -0.04] |
| d | [-1.10, -0.80] | [0.49, 0.96] | [-1.63, -1.25] | [-1.40, -0.90] | [0.20, 0.32] | [-0.26, -0.14] |
| e | [-1.30, -1.00] | [0.73, 0.92] | [-1.36, -1.01] | [-1.50, -1.40] | [0.37, 0.58] | [-0.36, -0.26] |

* Thresholds and parameter estimates are presented as [range], where the range represents the minimum and maximum values across repeated simulations.

**Table 2**

**Thresholds and parameter estimates under the pooling approach.**

| | Wiedemann model | | | ML model | | |
|---|---|---|---|---|---|---|
| Intersection No. | Thresholds | $\sigma$ | $\xi$ | Thresholds | $\sigma$ | $\xi$ |
| a | -0.80 | 1.26 | -2.76 | -1.60 | 0.58 | -0.34 |
| b | -1.40 | 2.90 | -2.07 | -1.40 | 0.40 | -0.28 |
| c | -1.00 | 0.51 | -0.82 | -1.50 | 0.42 | -0.33 |
| d | -0.80 | 0.55 | -1.52 | -0.80 | 0.21 | -0.15 |
| e | -1.10 | 0.59 | -1.01 | -1.40 | 0.41 | -0.28 |

Subsequently, annual crash frequencies were predicted for each intersection under both the pooling and non-pooling approaches for the Wiedemann and ML models. Under the pooling approach, annual crash frequencies for each intersection were directly predicted using Eqs. (7) and (8). Under the non-pooling approach, crash frequencies were first estimated for each simulation repetition using Eqs. (7) and (8), and were then averaged across repetitions to obtain an intersection-level estimate. In addition, a separate crash frequency prediction for each intersection was made directly from the crashes generated by the ML model, by applying Eq. (8) to the ML-simulated hourly crash frequency. Finally, the prediction accuracy of each crash frequency prediction approach was evaluated across all intersections using the hit rate, MAE, and RMSE.

**Table 3**
**Results of crash frequency prediction.**

| | Predicted crash frequency | | | | | Ground truth | | |
|---|---|---|---|---|---|---|---|---|
| | Non-pooling | | Pooling | | | | | |
| Intersection No. | From Wiedemann conflicts | From ML conflicts | From Wiedemann conflicts | From ML conflicts | From ML crashes | Actual crash frequency | Poisson estimator Lower | Poisson estimator Upper |
| a | (0.00) | **1.09** | (0.00) | **0.72** | 0 | 1 | 0.21 | 2.92 |
| b | (0.00) | **1.79** | (0.00) | 0.01 | 0 | 1 | 0.21 | 2.92 |
| c | 0.00 | **0.24** | 0.00 | 0.00 | 1752 | 0.67 | 0.08 | 2.41 |
| d | (0.00) | **4.25** | (0.00) | **2.11** | 47304 | 2.67 | 1.15 | 5.25 |
| e | (0.00) | **0.49** | (0.00) | **0.04** | 0 | 0.33 | 0.01 | 1.86 |
| Hit rate | 0% | ***100%*** | 0% | 60% | 0% | | | |
| MAE | 0.67 | 0.61 | 0.67 | ***0.56*** | 9811 | | | |
| RMSE | 0.67 | 0.82 | 0.67 | ***0.62*** | 21168 | | | |

* Notes: Numbers in parentheses: estimates that are invalid due to $\xi$ < -1.0; Bold: predicted crash frequency falls within the 95% Poisson confidence interval for the actual crash frequency; Bold and italics: best performance among all models for each metric.

Table 3 summarizes the comparison results, showing the predicted annual crash frequency, the actual crash frequency, and the corresponding 95% Poisson confidence intervals for each intersection. Note that according to Smith (1985), results are considered unreliable or invalid when the estimated shape parameters $\xi$ (as reported in Tables 1 and 2) are less than -1.0. Here, we obtained such $\xi$ values when fitting EVT models to the conflicts generated by the Wiedemann model for intersections a, b, d, and e, regardless of whether a pooling or non-pooling approach was used. As shown in Fig. 5, this might be due to the discontinuity and multi-modality of the minimum TTC values generated by the Wiedemann model. In line with previous EVT-based crash prediction literature (Zheng et al., 2014) we excluded these predictions from the model comparison, but it should be noted that regardless of whether we include them or not, this does not change the overall conclusions.

With regards to the hit rate, the ML conflict-based predictions consistently outperformed the Wiedemann conflict-based predictions under both pooling and non-pooling approaches. Specifically, for the non-pooling approach, the ML model achieved a 100% hit rate (5 out of 5), higher than the 0% (0 out of 5) observed for the Wiedemann model. Similarly, in the pooling approach, the ML model maintained a higher hit rate of 60% (3 out of 5) compared to the 0% (0 out of 5) for the Wiedemann model. As for the crash frequency predictions made directly from the crashes generated by the ML, these were either zero or much too high, and therefore never fell within the 95% Poisson confidence intervals.

As for the MAE and RMSE metrics, again the ML conflict-based crash prediction achieved the best results, exhibiting the lowest MAE and RMSE values regardless of the pooling approach, with lowest error values for the pooling approach. This is followed by the Wiedemann conflict-based crash prediction, while the ML crash-based crash prediction shows the highest values. These results indicate that the ML conflict-based approach outperforms the Wiedemann conflict-based and ML crash-based approaches in terms of crash frequency prediction accuracy, with the latter showing the weakest performance overall.

## 4. Discussion

### *4.1. Main findings*

The comparison of the Wiedemann vs ML-based crash frequency predictions yielded three major findings: First, the ML model exhibited driving behaviours that seemed generally closer to that of human drivers than the Wiedemann model. The ML model was trained on real-world data that capture a wide range of driving behaviours exhibited by human drivers across diverse traffic scenarios (Lioutas et al., 2025). The results presented here suggest that this exposure allowed the model to learn and reproduce driving patterns that resemble human behaviour. In contrast, the Wiedemann model, which relies on predefined driving rules, lacks the flexibility to capture the variability and subtleties inherent in human driving, resulting in less human-like performance across the evaluated metrics.

Second, the ML-based conflict model consistently achieved higher crash frequency prediction accuracy than the Wiedemann-based conflict model across both pooling and non-pooling approaches, comprehensively confirming our first main hypothesis. The human-likeness of the ML model behaviour, and in particular the more human-like TTC distributions, suggest that the model behaviour has learned to enter into and resolve conflicts in a similar way to humans. Again, this is presumably because the real-world ML model training datasets contain a large number of conflicts. For this reason, the ML model may be capable of generating conflicts in a given infrastructure rather well, better than the rule-based Wiedemann model, which in turn allows a more accurate prediction of crashes from the ML model conflicts.

In fact, the Wiedemann model did not only provide less accurate crash frequency predictions, it generally failed to generate any meaningful predictions at all. This may seem surprising given that, as mentioned above, other authors have successfully used the Wiedemann model together with EVT to predict crash frequencies (Wang et al., 2018). However, it should be noted that all of this past work included a step where the Wiedemann model was directly calibrated to trajectory data from the studied intersections. The use here of default parameter values for the Wiedemann model may have caused its irregular, multimodal low-TTC distributions (Fig. 5), in turn yielding invalid or zero EVT predictions. In the present study, the aim was precisely to investigate the possibility of generating crash frequency predictions *without* any model calibration data, because acquiring such data requires first building the road infrastructure in question, thus invalidating a key benefit of traffic microsimulation as a proactive, pre-implementation tool for infrastructure design.

A reasonable question here is whether other rule-based models besides the Wiedemann model would fare better in this comparison. As mentioned above in this paper, the Wiedemann model was adopted here because of its use in past EVT based crash prediction research (Wang et al., 2018). Some past work simulating conflicts for road safety evaluation (Qi & Zheng, 2025) has instead used the Intelligent Driver Model (IDM) (Treiber et al., 2000). Therefore, to test whether the poor results for the Wiedemann model were due to shortcomings of this specific model in the intersections studied here, the same simulations and EVT analyses were rerun, instead using the IDM with its default parameters in SUMO. The results of this analysis were very similar to the results for the Wiedemann model, showing similar problems with multimodal low-TTC distributions (Fig. 5), and again with all EVT crash frequency predictions being invalid or zero. This result provides further evidence in favour of our first main hypothesis, regarding the benefit of ML-based over rule-based traffic microsimulation in this context.

The third and final main finding was that, while crashes occurred in the ML-based microsimulation, directly using these simulated crashes to predict real-world crash frequency achieved the poorest prediction performance. This confirmed our second main hypothesis. As suggested in the Introduction, this is likely due to the scarcity of crashes in the training data (Liu & Feng, 2024; Yan et al., 2023). Due to this data scarcity, the ML model has not been trained to reproduce human behaviour patterns that cause crashes, which as seen in Table 3 can result in simulated crashes occurring at a frequency that is sometimes zero and sometimes many orders of magnitude larger than human crash frequencies.

It should be noted that no crashes were observed in the Wiedemann or IDM simulations, as the model's rule-based structure inherently maintains safe distances and crash-free driving (Hamdar & Mahmassani, 2008; He et al., 2013; Van Lint & Calvert, 2018).

### *4.2. Limitations*

This study has a few limitations that can be addressed in future research. First, this study compared only a single ML model with the traditional Wiedemann model and IDM; it will be useful to investigate a wider range of alternative ML models. Second, existing research has used traffic conflicts combined with EVT to predict crash frequencies for different crash types (e.g., rear-end, lane-change, and crossing crashes; (Wang et al., 2018), whereas this study did not separate different crash types,  due to the lack of crash type information in the available crash statistics dataset. Third, previous studies have applied traffic conflicts in combination with EVT to predict crash frequencies for different severity levels (e.g., severe and non-severe crashes; (Hussain et al., 2024). However, our crash dataset was not large enough to permit such stratified analysis, which would also have required use of additional surrogate safety measures and more complex, multivariate EVT models (Hussain et al., 2024). Fourth, as this study focused solely on signalized intersections, the ML model's ability to predict crash frequency has not yet been examined in unsignalized intersections. Future research could address these limitations by incorporating multiple ML models, extending the analysis to different crash types and crash severity levels using simulated conflicts, and evaluating the proposed framework in other intersection types.

### *4.3. Outlook: Using ML models for crash frequency prediction*

As mentioned in the Introduction, existing ML-based microsimulation frameworks have been primarily developed for testing and validating AVs, rather than being explicitly designed for crash frequency prediction. From the application, here, of ML-based microsimulations to traffic safety assessment, several insights emerge for how ML models of road user behaviour could be further developed to better support this application of the models.

First, traffic safety assessment typically requires traffic microsimulations at hourly scale. In contrast, most current ML-based microscopic simulators are only able to maintain reliable performance over short durations, typically ranging from a few seconds to a few minutes (Bhattacharyya et al., 2022; Tan et al., 2025). As a result, long-duration simulations must currently be decomposed into multiple short segments. Second, in real-world traffic systems, vehicles continuously enter and exit the network at designated locations. In contrast, ML-based microscopic simulators typically initialise vehicles in random configurations at the beginning of each simulation. Third, infrastructure safety evaluation, an example of traffic safety assessment, requires simulation of the models in the specific infrastructure design under study. However, most ML-based simulators are restricted to a limited set of pre-defined maps and scenarios developed during model training. Very few microscopic simulators, such as TorchDriveSim, currently allow the direct use of their built-in trained models on flexible and customised maps and scenarios.

The abovementioned requirements are all met by rule-based microscopic simulators, such as SUMO, which therefore remain dominant tools for traditional traffic safety analysis. Looking forward, there is a need for the development of ML-based microscopic simulators that can integrate long-duration microsimulation, continuous vehicle generation, and flexible scenario customisation. Such advancements would enable easier and more efficient implementation of ML-based crash frequency prediction studies.

## 5. Conclusion

This study employed state-of-the-art, data-driven ML-based traffic microsimulation to improve the accuracy of crash frequency prediction, using the Wiedemann as a benchmark for comparison. EVT was used to develop crash prediction models from simulated conflicts identified using two-dimensional TTC. The methodology was applied to five signalized intersections in Leeds, UK. The ML-based microsimulations enabled higher accuracy in crash frequency prediction from simulated conflicts. This better performance can be attributed to the ML model being trained on large quantities of real human driving data. However, directly using simulated crashes from the ML-based microsimulations to predict real-world crash frequency resulted in the lowest prediction accuracy among the approaches considered, presumably because the ML model's training data had few or no crashes. Overall, the methodology proposed in this study demonstrates the strong potential of ML models for crash frequency prediction from simulated conflicts, supporting the development of more reliable conflict-based safety analysis using traffic microsimulation. To make future application of these promising methods more efficient, improved ML-based traffic microscopic simulators could be developed, integrating long-duration simulation with continuous vehicle entry/exit in arbitrary infrastructure designs.

## Acknowledgements

The authors would like to thank Christopher Way and Dylan Turner at Leeds City Council and Neil Hudson at West Yorkshire Combined Authority, for sharing data on crash statistics and traffic flow data.